%% file: samplepaper.tex
\documentclass[runningheads]{llncs}
\usepackage[T1]{fontenc}
\usepackage{graphicx}
\usepackage{bbm}
\usepackage{subcaption}
\usepackage{multirow}
\usepackage{algorithm}
\usepackage{algpseudocode}
\usepackage{amsmath}
\usepackage{hyperref}
\usepackage{amssymb}

\begin{document}
\title{SemAlign: Language Guided Semi-supervised Domain Generalization}
%
%
\author{Muditha Fernando\inst{1} \and
Kajhanan Kailainathan\inst{1} \and
Krishnakanth Nagaratnam\inst{1} \and Isuranga Udaravi Bandara Senavirathne\inst{1} \and Ranga Rodrigo\inst{1}}
%

%
\institute{University of Moratuwa}
\maketitle              
\input{sec/0_abstract}
\input{sec/1_introduction}
\input{sec/2_relatedWork}
\input{sec/3_method}
\input{sec/4_results}
\input{sec/5_conclusion}
%
%
%
%

\end{document}

%% file: sec/0_abstract.tex
\begin{abstract}
Semi-supervised Domain Generalization (SSDG) addresses the challenge of generalizing to unseen target domains with limited labeled data. Existing SSDG methods highlight the importance of achieving high pseudo-labeling (PL) accuracy and preventing model overfitting as the main challenges in SSDG. In this light, we show that the SSDG literature's excessive focus on PL accuracy, without consideration for maximum data utilization during training, limits potential performance improvements.
We propose a novel approach to the SSDG problem by aligning the intermediate features of our model with the semantically rich and generalized feature space of a Vision Language Model (VLM) in a way that promotes domain-invariance. The above approach is enhanced with effective image-level augmentation and output-level regularization strategies to improve data utilization and minimize overfitting. Extensive experimentation across four benchmarks against existing SSDG baselines suggests that our method achieves SOTA results both qualitatively and quantitatively. The code will be made publicly available.

\keywords{Semi-supervised Domain Generalization  \and Vision Language Models.}
\end{abstract}

%% file: sec/1_introduction.tex
\section{Introduction}
\label{sec:intro}

The semi-supervised domain generalization (SSDG) problem setting jointly addresses the constrains of the domain generalization (DG) and semi-supervised learning (SSL) problems. DG aims to generalize well to unseen target domains \cite{DGtheory,metaDG,alignment1}, while SSL aims to maximize results with limited labeled data and an abundance of unlabeled data \cite{sslentropy,sslpseudolabel,fixmatch,sslmeanteacher}. In the SSDG setting, the performances of state of the art (SOTA) DG and SSL algorithms are suboptimal. SSL methods such as FixMatch \cite{fixmatch} demonstrate relatively better performance than DG methods in the SSDG setting. Hence, the majority of SSDG methods build on top of FixMatch with the objective of improving model-generalization \cite{stylematch,multimatch,FBCSA,dgwm}.

\begin{figure}[t]
  \centering
   \includegraphics[width=0.6\linewidth]{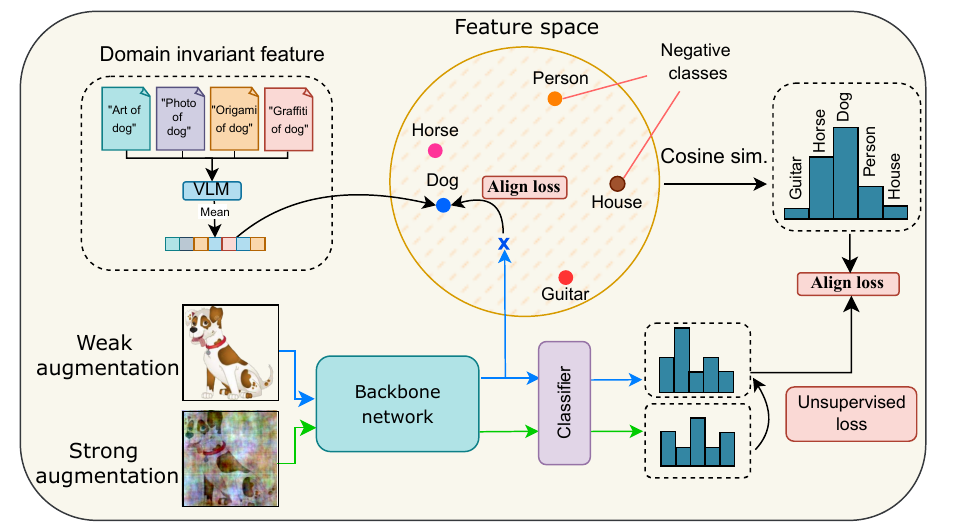}
   \caption{A conceptual diagram illustrating the proposed domain-invariant semantic alignment approach, where features are aligned with semantically rich representations from the text encoder of a VLM to achieve semi-supervised domain generalization. }
   \label{fig:teaser}
\end{figure}

\begin{figure*}[t]
    \centering
    \includegraphics[width=1\textwidth]{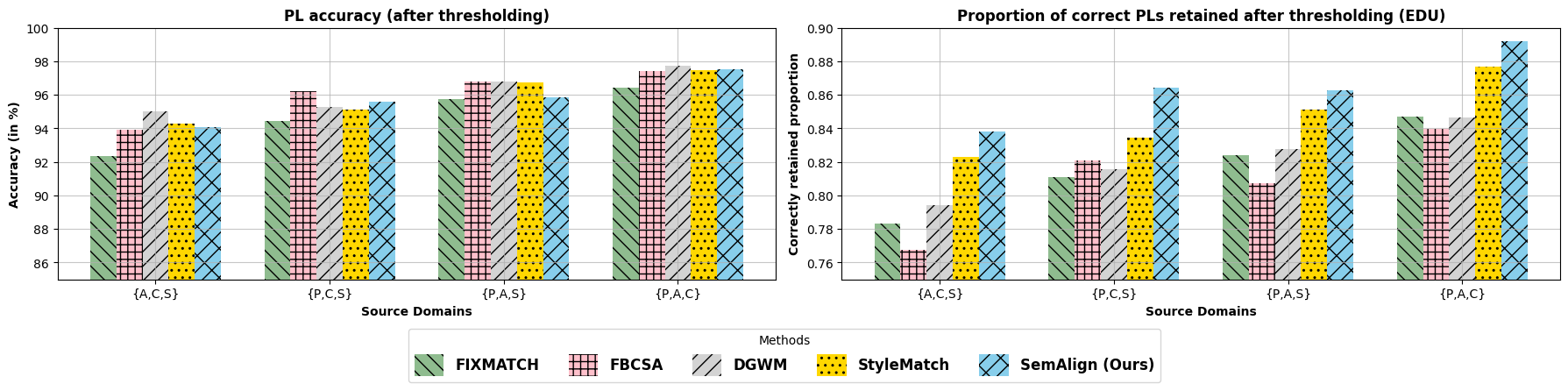} 
    \caption{Left: PL accuracy after thresholding, Right: The correct pseudo-labels retained after thresholding as a proportion of the entire dataset (EDU) of the baselines FixMatch ~\cite{fixmatch}, FBC-SA ~\cite{FBCSA}, DGWM ~\cite{dgwm}, StyleMatch ~\cite{stylematch}, and ours. Our method achieves significant gains in proportion of samples retained after thresholding while maintaining comparable PL accuracy to SOTA SSL-based-SSDG methods. Here A, C, P, and S
denote Art-painting, Cartoon, Photos, and Sketch domains, respectively.}
    \label{fig:PL_Stats}
\end{figure*}

Although these methods have significantly improved upon FixMatch, room for improvement remains. Commonly cited challenges in the SSDG literature are improving pseudo-labeling accuracy and minimizing overfitting to limited labeled data ~\cite{FBCSA,stylematch}. Obtaining accurate pseudo-labels (PL) is especially    challenging under multiple domain shifts ~\cite{stylematch,dgwm}. Therefore, restricting domain-specific feature learning and promoting domain-invariant feature learning is crucial. Further, an analysis of PL statistics in SSL-based SSDG methods reveals that, for several methods, the total number of correct PLs retained after thresholding falls even below that of the SSL baseline~\cite{fixmatch}, as shown in Fig.~\ref{fig:PL_Stats}. We term this metric Effective Data Utilization (EDU). EDU quantifies the number of samples effectively contributing to model training. This metric is critical, as a low EDU suggests that improvements in PL accuracy may merely be an artefact of overfitting to a smaller, easier subset of data. 
In addition to correct PLs, valuable information exists in rejected samples as well. The inability of current SSDG methods to leverage them is also a key limitation.

We propose a new SSDG approach towards addressing the above limitations from three perspectives: feature space, model output space, and input/pixel space. 
The feature space of the model is leveraged to tackle the pseudo-labeling accuracy problem by aligning visual features with class-label semantic priors from a pre-trained language encoder. This allows the model to map the visual features to a rich semantic space ~\cite{devise,allLabels}.  We use the CLIP text encoder to generate these priors, using a technique that enables these class features to be domain-invariant ~\cite{sentence}. 
CLIP has been trained on a large-scale, diverse dataset, giving it the capability to generalize across domains ~\cite{practicaldg,sentence}. Also, the CLIP text encoder has a robust understanding of image features, since it has been aligned with the CLIP image encoder ~\cite{textdiff}.  
Therefore, we utilize the label embedding space from the CLIP text encoder and refine it following Path-CLIP ~\cite{pathclip}, to strike a balance between task-specific adaptation and preserving CLIP's rich visual understanding.
We further regularize the feature space alignment by contrasting with negative class label embeddings.
We also attend to the model output space to address some of the aforementioned limitations. We introduce Entropy Meaning Loss (EML) following FullMatch ~\cite{fullmatch}, which improves PL accuracy by reducing the competitiveness of negative classes. This also improves the EDU by enabling more samples to pass the thresholding criterion ~\cite{fullmatch}.
To incorporate samples rejected at thresholding to the learning process, we introduce Adaptive Negative Learning (ANL) following FullMatch ~\cite{fullmatch}. ANL 
allows the model to leverage all unlabeled data for training.
To tackle the overfitting issue we follow StyleMatch ~\cite{stylematch} and use a stochastic classifier ~\cite{SNN,snn_bayesian}. This prevents the model from producing excessive incorrect PLs. To facilitate domain invariant feature learning, StyleMatch ~\cite{stylematch} uses style-transferred images from multiple domains. However, this adds a large overhead at training time (Tab. ~\ref{tab:aug_overhead}). Instead, in pixel space we introduce a set of specialized data augmentations such as Fourier-based augmentations ~\cite{fourierDG,FourierSSDG} and textural feature reduction. By these augmentations, we remove the domain-specific components of the input data, thus forcing the model to learn the semantic features instead of relying on spurious correlations, with a significantly lower overhead.

\vspace{0.1cm}
\noindent
In summary, we make the following key contributions: 

\begin{enumerate}
    \item To the best of our knowledge, we are the first to leverage the rich output space of a Vision Language Model (\textbf{VLM}) as domain-invariant class priors to enhance pseudo-labeling accuracy in SSDG. 
    \item We highlight the importance of considering both the pseudo-labeling accuracy and the \textbf{Effective Data Utilization (EDU)} in SSDG and show significant gains in the latter.
    \item We propose a novel approach to address the SSDG problem from three perspectives: (1) in the \textbf{feature space}, through alignment strategies; (2) in the model \textbf{output space}; and (3) in the \textbf{pixel space}, leveraging innovative data augmentation techniques. Our approach \textbf{does not rely on domain labels}. 
    \item We conduct experiments on four different DG datasets: PACS ~\cite{datasetPACS}, OfficeHome ~\cite{datasetOfficeHome}, DigitsDG ~\cite{datasetDigitsDG}, and miniDomainNet ~\cite{ens1}, a subset of DomainNet ~\cite{domainNet}, demonstrating \textbf{state-of-the-art (SOTA) performance} compared to other SSDG methods in the literature.
\end{enumerate}

%% file: sec/2_relatedWork.tex
\section{Related Work}
\label{sec:relatedwork}

Stylematch ~\cite{stylematch} introduces the SSDG problem by pointing out the shortcomings when DG and SSL methods are naively adopted to the SSDG setting. One of the main issues addressed by most SSDG methods is the substantial drop in PL accuracy when multiple domains are found in training data ~\cite{stylematch,FBCSA,multimatch,dgwm}. One approach in SSDG exploits domain-specific features in the input to improve PL accuracy ~\cite{dgwm,multimatch}. DGWM ~\cite{dgwm} learns a domain guided weight masking algorithm to achieve this, while MultiMatch ~\cite{multimatch} uses a criterion to fuse results from multiple domain-specialized classifiers and a global classifier.
A second approach uses a shared classifier similar to the ERM baseline ~\cite{ERM} where the penultimate features are expected to be domain invariant.
StyleMatch replaces the classifier in FixMatch with a stochastic classifier ~\cite{SNN} to minimize model overfitting, and a multiview consistency framework is used to encourage the feature encoder outputs to be domain-invariant.
UPLM ~\cite{SSDG_Uncertainty} improves PL accuracy by using an uncertainty parameter to filter out uncertain PLs. FBC-SA ~\cite{FBCSA} leverages domain-aware prototypes derived from the labeled data to generate posteriors from the feature space, which are then aligned with the model output space. However, domain alignment is used to subsequently promote domain invariance in the feature space as training progresses. Our method too follows this second approach. 
Unlike the above methods, we utilize all three of input pixel space, feature space, and output space to improve PL accuracy and promote domain invariance. We also learn from all the samples and show its effectiveness in the SSDG setting. 



%% file: sec/3_method.tex
\section{Method}

\begin{figure*}[t]
    \centering
    \includegraphics[width=\textwidth]{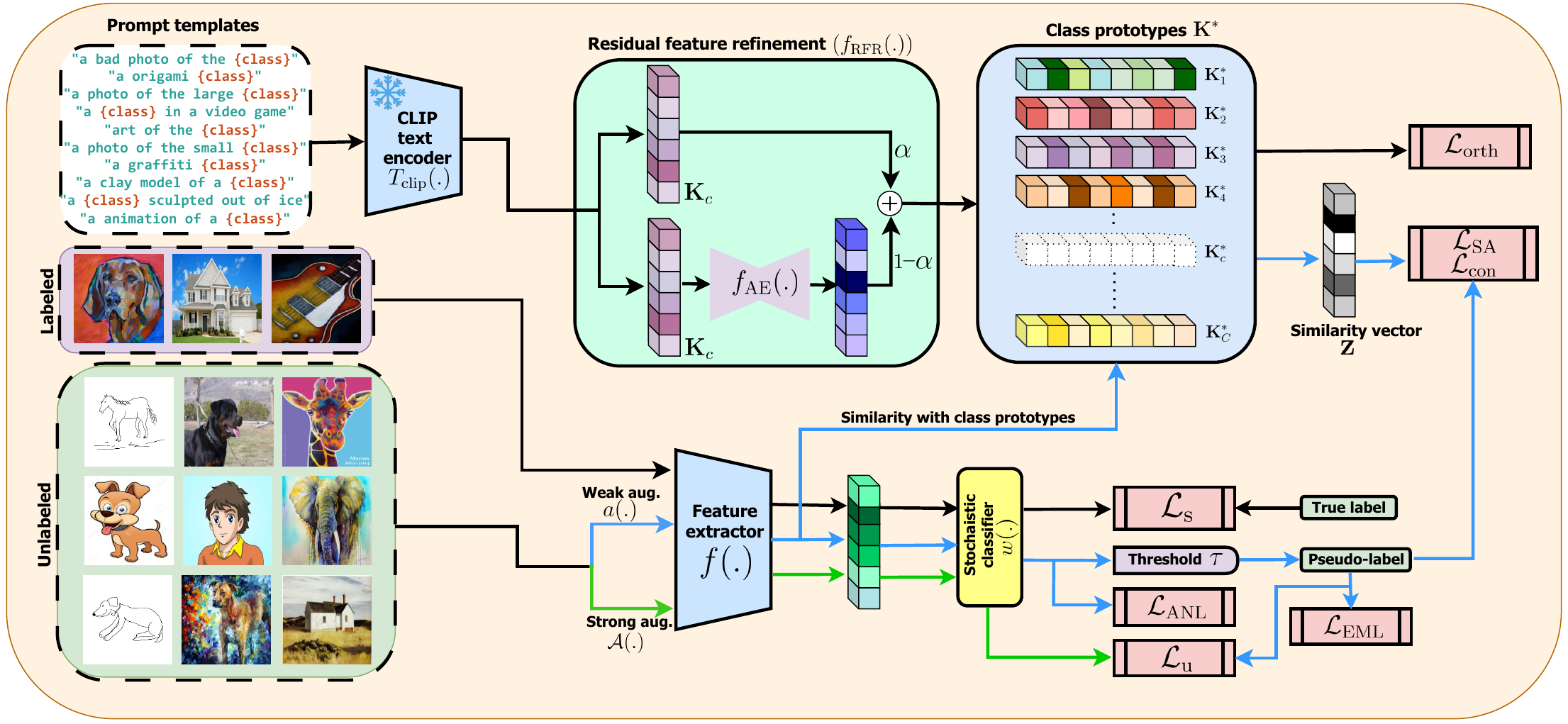} 
    \caption{For class labels of the classification problem we generate domain invariant representations using CLIP \cite{CLIP} text encoder and further refine them using the RFR module to create the class prototype matrix ($\mathbf{K}^*$). The orthogonality of these class prototypes is encouraged using $\mathcal{L}_\mathrm{orth}$. 
    For a weakly augmented view, we align the feature extractor output with the corresponding class prototype guided by the PL generated by the classifier. Alignment is constrained using $\mathcal{L}_\mathrm{SA}$ and $\mathcal{L}_\mathrm{con}$. To reduce overfitting we use a stochastic classifier. At the classifier level, we minimize competition from competitive classes using $\mathcal{L}_\mathrm{EML}$ and allow learning from rejected samples at the pseudo-labeling using $\mathcal{L}_\mathrm{ANL}$. $\mathcal{L}_\mathrm{s}$ and $\mathcal{L}_\mathrm{u}$ are same as in FixMatch \cite{fixmatch}.}
    \label{fig:Architecture}
\end{figure*}

Our SemAlign architecture includes a pre-trained text encoder, a Residual Feature Refinement (RFR) module, an image feature extractor, and a stochastic classifier. The text encoder generates domain-invariant class prototypes, and the feature extractor outputs are aligned with these prototypes. Further, objectives are used to encourage more PLs to clear the thresholding criterion and to learn from rejected samples. Enforcing consistency between weakly and strongly augmented image views further promotes domain-invariant learning. Figure \ref{fig:Architecture} provides an overview of our method.
The algorithm is provided in the suppl.

\subsection{Problem Setting} 
In our work, $\mathcal{X}$ is the input image feature space and $\mathcal{Y} = \{1, 2, ..., C\}$ is the output space, a finite set of $C$ possible classes.
A domain is defined as a joint probability distribution $P_{XY}$ over $\mathcal{X} \times \mathcal{Y}$. During training time, we will have access to $D$ source datasets $\mathcal{\{S}^{(d)}\}_{d=1}^D$, sampled from related but distinct joint probability distributions, $\{P_{XY}^{(d)}\}_{d=1}^D$, each representing a distinct domain. Note that $\forall d, d'\in\{1, 2, ..., D\}, d \neq d' \Rightarrow P_{XY}^{(d)} \neq P_{XY}^{(d')}$. Each source dataset $\mathcal{S}^{(d)}$ consists of labeled and unlabeled data, denoted $\mathcal{S}_{\ell}^{(d)}$ and $\mathcal{S}_{u}^{(d)}$ respectively. That is, $\mathcal{S}^{(d)}$ = $\mathcal{S}_\ell^{(d)} \cup \mathcal{S}_u^{(d)}$, with $\mathcal{S}_\ell^{(d)} = \{(\mathbf{x}_i^{(d)}, y_i^{(d)})\}\sim_{i.i.d}P_{XY}^{(d)}$ and $\mathcal{S}_u^{(d)} = \{(\mathbf{x}_i^{(d)})\} \sim_{i.i.d} P_{X}^{(d)}$, where $P_{X}^{(d)}$ is the marginal distribution of $P_{XY}^{(d)}$ over $\mathcal{X}$. The source datasets contain $n_\ell$ labeled samples and $n_u$ unlabeled samples. Generally, $n_\ell\ll n_u$ in SSDG (and SSL) settings. Each minibatch for training will comprise of $B_\ell$ labeled samples and $B_u$ unlabeled samples. Our objective is to learn a model $\mathcal{F}$ using the source datasets, such that $\mathcal{F}(\mathbf{x}_i^{(\mathcal{T})}) = y_i^{(\mathcal{T})}$ with high probability, generalizing well to an unseen target domain $P_{XY}^\mathcal{T} = \{(\mathbf{x}_i^{(\mathcal{T})}, y_i^{(\mathcal{T})})\}$. Here, $\forall d, P_{XY}^{(\mathcal{T})} \neq P_{XY}^{(d)}$. We decompose $\mathcal{F} = w\circ f$, where $f:\mathbf{x}\rightarrow \mathbf{h}$ is a feature extractor and $w:\mathbf{h}\rightarrow y$ is a classifier. We also note that, for our implementation, meaningful class label names are required, while domain labels are not required.


\subsection{The Preliminary Framework}
Since the FixMatch framework \cite{fixmatch} performs better than DG methods and other SSL methods in the SSDG setting \cite{stylematch}, we build our method on top of FixMatch. Given a batch of labeled and unlabeled images, FixMatch first applies weak augmentation ($a(.)$) (e.g., random flips and shifts) to the labeled samples and computes the standard cross-entropy loss ($\mathcal{L}_s$) in a supervised manner. 
\begin{equation}
    \label{eq:supervisedLoss}
    \mathcal{L}_s = \frac{1}{B_\ell}\sum_{b=1}^{B\ell}H(P_b\ ,Q_b^w),
\end{equation}
where $P_b$ is the one-hot distribution of the given label, and $Q_b^w = P(y\ |\ w,f(a(\mathbf{x}_b))))$ is the model probability output distribution for input $\mathbf{x}_b$.
For unlabeled data, it generates PLs using the model's predictions on weakly augmented views, retaining only those with high confidence. These PLs are then used as targets for images after applying strong augmentation ($\mathcal{A(.)}$) (e.g., RandAugment \cite{randaugment}), enforcing consistency through a cross-entropy loss.
\begin{equation}
    \label{eq:unsupervisedLoss}
    \mathcal{L}_u = \frac{1}{B_u} \sum_{b=1}^{B_u} \mathbbm{1}(\max(Q_b^w) \geq \tau) H(\hat{Q}_b^w,\ P(y \mid w, f(\mathcal{A}(\mathbf{x}_b)))),
\end{equation}
where $\tau$ is a scalar hyperparameter denoting the threshold above which we retain a PL and $\hat{Q}_b^w = \mathrm{arg\ max}(Q_b^w)$ is the PL. For simplicity, we assume that $\mathrm{arg\ max}$ applied to a probability distribution produces a corresponding ``one-hot" distribution.
The overall objective is a combination of the supervised and unsupervised losses.

Following StyleMatch \cite{stylematch}, we replace the original classifier in the FixMatch framework with a stochastic classifier \cite{SNN}, to improve the generalization capability of the model by minimizing overfitting to limited labeled data. 


\subsection{Aligning with CLIP Text Embeddings} \label{clip_align}

We incorporate additional domain-invariant and semantically rich class prior information from the CLIP text encoder \cite{CLIP} to guide the training process.

\vspace{0.1cm}
\noindent
\textbf{Domain-Invariant Class Prototypes: }To utilize prior information from the CLIP text encoder, we create $C$ domain-invariant class prototypes $\{\mathbf{K}_c\}_{\mathbf{c}=1}^C$ corresponding to the $C$ classes of the classification problem, at the beginning of training. 
$\mathbf{K}_c$ is generated by inputting the corresponding class label c into multiple prompt templates, encoding them with the frozen pre-trained CLIP text encoder, and subsequently averaging the resulting embeddings (see Fig. \ref{fig:Architecture}), inspired by RISE \cite{sentence}.
We use a subset of the recommended list of eighty templates of text prompts by CLIP \cite{CLIP,sentence} to span the entire meta-domain of images. 
Empirical results show that this approach is superior to using a single template \cite{sentence}.

\vspace{0.1cm}
\noindent
\textbf{Residual Feature Refinement: }
The RFR module is used to mitigate the catastrophic fogetting of previously 
learned knowledge while optimizing an orthogonality loss term ($\mathcal{L}_\mathrm{orth}$) that encourages class features to become mutually orthogonal, promoting adaptation to classification tasks. The inspiration for the RFR module is taken from Path-CLIP \cite{pathclip}, however, we use it for a different objective. 
\begin{equation}
    \label{eq:rfr}
    \mathbf{K}_c^* = f_\mathrm{RFR}(\mathbf{K}_c) = (1-\alpha)\cdot f_\mathrm{AE}(\mathbf{K}_c) + \alpha \mathbf{K}_c, 
\end{equation}
where $f_\mathrm{RFR}(.)$ describes the RFR module, $\mathbf{K}_c^*$ is the refined class prototype where prototypes of distinct but potentially similar classes are encouraged to be orthogonal, $f_\mathrm{AE}(.)$ is a 2-layer fully connected auto-encoder module, and the hyperparameter $\alpha$ is the ratio between preserving original knowledge from CLIP and adapting to fine-tuned knowledge. $\alpha$ is typically a high value (e.g., 0.9). The orthogonality loss is calculated as follows:
\begin{equation}
    \label{eq:orthogonalityLoss}
    \mathcal{L}_{\mathrm{orth}} = \frac{1}{C^2-C}||\mathbf{K}^* \mathbf{K}^* \mathbf{^T-I_{C \times C}}||_F^2,
\end{equation}
where $\mathbf{K}^* = [\mathbf{K}_1^*;\mathbf{K}_2^*;...;\mathbf{K}_C^*]$ is the class prototype matrix, $||.||_F^2$ is the Frobenius norm. 
This helps us to better distinguish similar and confusing classes, as shown in Sec. \ref{ablation_impact}.  

\vspace{0.1cm}
\noindent
\textbf{Feature Similarity and Alignment: }
We compute posterior class similarity scores for each weakly augmented image feature embedding using domain-invariant class priors from the CLIP text encoder, following the same approach as in CLIP zero-shot classification. The similarity vector $\mathbf{z} \in \mathbb{R}^C$ is computed for input $\mathbf{x}_b$ as follows:
\begin{equation}
    \label{eq:similarityVector}
    \mathbf{z} =  \mathbf{K}^{* \mathbf{T}}f(a(\mathbf{x}_b)) ,
\end{equation}
By computing $\mathrm{softmax}(\mathbf{z})$, we obtain the semantic class posterior probability distribution, $Q_b^{\mathrm{sem}} = P_b(y\ |\ \mathbf{K}^*,f(a(\mathbf{x}_b)))$. This distribution is used to align the class posterior probabilities $Q_b^w = P_b(y\ |\ w, f(a(\mathbf{x})))$ from the model output space. The alignment is done by minimizing a cross-entropy loss:
\begin{equation}
    \label{eq:wordAlignLoss}
    \mathcal{L}_\mathrm{SA} = \frac{1}{B_u}\sum_{b=1}^{B_u} \mathbbm{1}(\max(Q_b^w) \geq \tau) H(Q_b^{\mathrm{sem}}\ |\ Q_b^w) ,
\end{equation}
This alignment with semantic and domain-invariant class representation $\mathbf{K}^*$ enhances the representation quality of the image feature representation $f(a(\mathbf{x}_b))$.


\vspace{0.1cm}
\noindent
\textbf{Improving Feature Discrimination: }
Taking inspiration from FBC-SA \cite{FBCSA}, we resort to sharpening the posterior class similarity scores $\mathbf{z}$, by increasing the contrast of the similarity of the predicted class, against the similarity of other classes. The following loss function ($\mathcal{L}_\mathrm{con}$) is used to achieve this objective.
\begin{equation}
    \label{eq:contrastiveLoss}
    \mathcal{L}_{\mathrm{con}} = 1-\mathbf{z}[y_\mathrm{pred}] + \frac{1}{C-1}\sum_{y_\mathrm{c} \neq y_\mathrm{pred}}\mathbf{z}[y_\mathrm{c}],
\end{equation}
where $y_\mathrm{pred}$ is the model prediction of the weakly augmented view, $w(f(a(\mathbf{x}_b)))$. 

We define the feature-level constraint $\mathcal{L}_\mathrm{feat}$ as the summation of the three complementary losses $\mathcal{L}_\mathrm{SA}$, $\mathcal{L}_\mathrm{orth}$, and $\mathcal{L}_\mathrm{con}$.
\begin{equation}
    \mathcal{L}_\mathrm{feat} = \mathcal{L}_\mathrm{SA} + \mathcal{L}_\mathrm{orth} + 
    \mathcal{L}_\mathrm{con}
\end{equation}

\subsection{Maximizing Learning from Unlabeled Data} \label{full_match_method}
To simultaneously improve the quality of PLs and to utilize all the data samples in the training process, we adopt training losses such as EML ($\mathcal{L}_\mathrm{EML}$) and ANL ($\mathcal{L}_\mathrm{ANL}$) from FullMatch \cite{fullmatch}, a method that addresses the SSL problem. 

\vspace{0.1cm}
\noindent
\textbf{Entropy Meaning Loss: }
For the PLs that meet the thresholding criterion, EML reduces competition for the pseudo-label class from competing classes with significantly large probability values. This is done by neutralizing the competitiveness of these competing classes by enforcing a uniform distribution among these classes, which will improve the quality of PLs. This will increase the likelihood that the data samples will overcome the thresholding criterion \cite{fullmatch}.

\vspace{0.1cm}
\noindent
\textbf{Adaptive Negative Learning: }
ANL helps to learn useful information from all the data samples, even if they fail to pass the thresholding criterion. Even for ambiguous examples with multiple classes having competitive probability values, some classes will have extremely low probabilities. An adaptive $k$-value is calculated by ensuring that the top-$k$ error rate is close to zero, and a uniform distribution is enforced on the remaining classes, since the model is highly certain that these classes do not correspond to the given example. The ANL objective exploits this knowledge by learning it using a negative learning approach \cite{negativeL1,negativeL2}. This allows the model to learn from all the data samples.

We ablate and demonstrate that these objectives not only enhance performance but also increase the EDU (Fig. \ref{fig:PL_Stats}).

We define the output-level constraint $\mathcal{L}_\mathrm{out}$ as the summation of the two complementary losses $\mathcal{L}_\mathrm{EML}$ and $\mathcal{L}_\mathrm{ANL}$. (i.e.)   $\mathcal{L}_\mathrm{out} = \mathcal{L}_\mathrm{EML} + \mathcal{L}_\mathrm{ANL}$. 

Our overall loss has four different terms, the supervised and unsupervised losses from \cite{fixmatch}, feature-level, and output-level constraints, $\label{finalLoss}
    \mathcal{L} = \mathcal{L}_\mathrm{s}+\mathcal{L}_\mathrm{u}+\mathcal{L}_\mathrm{feat}+\mathcal{L}_\mathrm{out}$ .

\subsection{Data Augmentations} \label{data_augs}



\begin{figure}[t]
  \centering
   \includegraphics[width=0.7\linewidth]{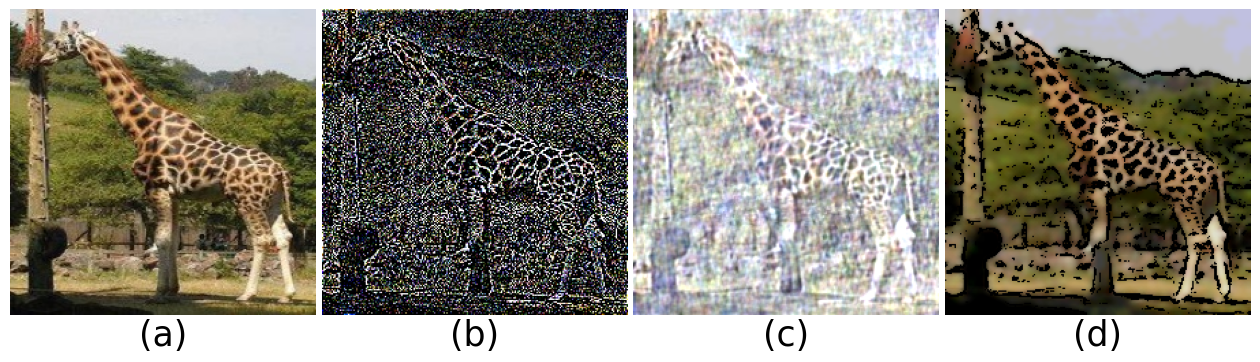}
   \caption{Our augmentation methods. (a) Raw image, (b) Phase-only image reconstruction, (c) Amplitude swapping, (d) Texture reduction. We apply these augmentations randomly along-side FixMatch \cite{fixmatch} augmentations as the strong augmentation.}
   \label{fig:augmentations}
\end{figure}


\vspace{0.1cm}
\noindent
\textbf{Augmentations in the Fourier Domain: }
Most semantic information resides in the phase component of a signal in the Fourier domain while the amplitude component primarily encodes non-semantic attributes such as texture and color \cite{phaseImportance}. We leverage this insight to generate data augmentations. While inspired by prior work \cite{fourierDG,FourierSSDG}, our approach differs in its methodology and application. We use two methods: (1) We retain only the phase component while setting the amplitude to a constant to preserve only the semantic information (see Fig. \ref{fig:augmentations} (b)). (2) We maintain a dynamically updated amplitude bank from previous samples and randomly swap the amplitude components of training samples with those from the bank to simulate novel visual domains and enhance generalization (see Fig. \ref{fig:augmentations} (c)).



\vspace{0.1cm}
\noindent
\textbf{Texture Reduction: }
CNNs are known to have a bias towards textural components in images over semantic and shape attributes \cite{textureBias}. This inherent bias affects model generalization. To mitigate it, we (1) reduce the image’s color space resolution to flatten nuanced variations and (2) apply smoothing using small kernels to distort textural features. However, this smoothing might blur the edges of the objects, distorting useful semantic information. To overcome this, we perform this smoothing after overlaying an edge mask of the image, which would prevent smoothing across these edges (see Fig. \ref{fig:augmentations} (d)).


%% file: sec/4_results.tex
\section{Experiments and Results}

\begin{table}[t]
    \centering
    
    \resizebox{\textwidth}{!}{%
    \begin{tabular}{@{}l|cccc|c||cccc|r@{}}
        \hline
        \textbf{\# labeled data}& \multicolumn{5}{|c||}{\textbf{5 labels}} & \multicolumn{5}{|c}{\textbf{10 labels}}\\
        \textbf{Model} & \textbf{PACS} & \textbf{OfficeHome}  & \textbf{DigitsDG} & \textbf{miniDomainNet} & \textbf{Avg} & \textbf{PACS} & \textbf{OfficeHome}  & \textbf{DigitsDG} & \textbf{miniDomainNet} & \textbf{Avg}\\
        \hline
        \textbf{FixMatch} \cite{fixmatch}  & 75.2 $\pm$ 1.3 & 55.2 $\pm$ 0.9 & 55.0 $\pm$ 2.8 & 34.6 $\pm$ 1.3 & 55.0 & 77.1 $\pm$ 2.0 & 58.2 $\pm$ 1.0 & 64.5 $\pm$ 1.4 & 39.8 $\pm$ 0.6 & 59.9 \\
        \textbf{StyleMatch} \cite{stylematch} & \underline{78.4 $\pm$ 0.9} & \underline{56.3 $\pm$ 0.6} & 55.9 $\pm$ 1.6 & \underline{36.6 $\pm$ 1.0} & 56.8 & \underline{79.7 $\pm$ 1.9} & \underline{59.8 $\pm$ 0.7} & 66.1 $\pm$ 1.6 & 40.6 $\pm$ 1.2 & 61.6 \\
        \textbf{FBC-SA} \cite{FBCSA} & 75.4 $\pm$ 1.1 & 55.8$\pm$ 0.7 & \underline{64.3 $\pm$ 0.5} & 36.1 $\pm$ 1.2 & \underline{57.9} & 78.1 $\pm$ 2.4 & 59.3 $\pm$ 1.0 & \underline{69.4 $\pm$ 1.4} & \underline{41.3 $\pm$ 1.1} & \underline{62.0} \\
        \textbf{DGWM} \cite{dgwm}& 78.0 $\pm$ 0.8 & \underline{56.3 $\pm$ 0.7} & 55.9 $\pm$ 0.6 & 36.4 $\pm$ 1.4 & 56.7 & 78.8 $\pm$ 1.6 & 59.6 $\pm$ 1.0 & 65.6 $\pm$ 1.5 & \underline{41.3 $\pm$ 1.5} & 61.3 \\
        \textbf{SemAlign (Ours)}  & \textbf{79.6 $\pm$ 0.5} & \textbf{57.0 $\pm$ 0.8} & \textbf{66.3 $\pm$ 0.6} & \textbf{36.7 $\pm$ 1.2} & \textbf{59.9} & \textbf{81.8 $\pm$ 1.7} & \textbf{60.3 $\pm$ 0.9} & \textbf{74.2 $\pm$ 1.4} & \textbf{42.2 $\pm$ 1.3} & \textbf{64.6} \\
        \hline
    \end{tabular}%
    }
    \caption{SSDG accuracy on target domain (\%). For each dataset, the average over 5 independent trials is reported. For each SSDG setting (5, 10 labels) the average over all datasets is reported. Best results are in \textbf{bold} and the second best is \underline{underlined}. We achieve an average performance gain of (+\textbf{2}\%)/(\textbf{1.68} times) and (+\textbf{2.6}\%)/(\textbf{2.23} times) in the 5 and 10 labels case, over the second best model.}
    \label{tab:ssdg_results}
\end{table}

\begin{table}[t]
    \centering
    \resizebox{0.6\columnwidth}{!}{%
    \begin{tabular}{@{}lr@{}}
        \hline
        Method & Average \\
        \hline
        Baseline \cite{fixmatch} & 77.1 \\
        Baseline + SC & 78.1 \\
        Baseline + SC + $\mathcal{L}_\mathrm{out}$ & 78.3 \\
        Baseline + SC + $\mathcal{L}_\mathrm{out}$ + Data aug  & 79.5 \\
        Baseline + SC + $\mathcal{L}_\mathrm{out}$ + Data aug + $\mathcal{L}_\mathrm{SA}$ & 80.8 \\
        Baseline + SC + $\mathcal{L}_\mathrm{out}$ + Data aug + $\mathcal{L}_\mathrm{SA}$ + $\mathcal{L}_\mathrm{con}$ & 81.1 \\
        Baseline + SC + $\mathcal{L}_\mathrm{out}$ + Data aug + $\mathcal{L}_\mathrm{SA}$ + $\mathcal{L}_\mathrm{orth}$ & 81.2 \\
        Baseline + SC + Data aug + $\mathcal{L}_\mathrm{SA}$ + $\mathcal{L}_\mathrm{orth}$ +  $\mathcal{L}_\mathrm{con}$& 81.0 \\
        Baseline + SC + $\mathcal{L}_\mathrm{out}$ + Data aug + $\mathcal{L}_\mathrm{SA}$ + $\mathcal{L}_\mathrm{orth}$ + $\mathcal{L}_\mathrm{con}$ (Ours) & \textbf{81.8} \\
        \hline
    \end{tabular}%
    }
    \caption{Ablation study on PACS with 10 labels per class. SC: stochastic classifier, $\mathcal{L}_\mathrm{out}$: Sec. \ref{full_match_method}, Data aug: Sec. \ref{data_augs}, $\mathcal{L}_\mathrm{SA}$, $\mathcal{L}_\mathrm{orth}$, $\mathcal{L}_\mathrm{con}$:  Sec. \ref{clip_align}).}
    \label{tab:ablation_study}
\end{table}

\begin{table}[t]
    \centering
    \caption{ Training overhead due to augmentations in comparison with FixMatch}
    \label{tab:aug_overhead}
    \resizebox{0.5\columnwidth}{!}{%
    \begin{tabular}{{@{}lcr@{}}}
        \hline
        Method &  Average time/epoch (sec)  & Overhead \\
        \hline
        FixMatch \cite{fixmatch} & 38.53 & -  \\
        StyleMatch \cite{stylematch}& 97.54 & 153.15\% \\
        SemAlign (Ours) & 60.2 & \textbf{56.24\%}\\
        \hline
    \end{tabular}%
    }
\end{table}

\vspace{0.1cm}
\noindent
\textbf{Datasets: }
We use the commonly used DG datasets, PACS \cite{datasetPACS}, OfficeHome \cite{datasetOfficeHome}, DigitsDG \cite{datasetDigitsDG}, and miniDomainNet, a subset of DomainNet \cite{domainNet} for our experiments. 

\vspace{0.1cm}
\noindent
\textbf{Training and Implementation: }
We follow the same training setup used in StyleMatch \cite{stylematch}. Experiments are done with 5 and 10 labeled data samples per class, and the labeled data is sourced randomly for each batch. A batch consists of 16 labeled and 16 unlabeled data samples. We use ResNet18 \cite{ResNeT} as our feature extractor with ImageNet \cite{imagenet} pretrained weights. The SGD optimizer is used and the feature extractor and the classifier head have initial learning rates 0.003 and 0.01 respectively, which eventually decay following the cosine annealing rule. We train all the models for 20 epochs on all datasets. We report top-1 accuracy averaged over 5 independent trials. 

\vspace{0.1cm}
\noindent
\textbf{Evaluation Protocol: }
We use the leave-one-domain out protocol \cite{domainbed} that has been adopted by most prior work in SSDG \cite{stylematch,FBCSA,dgwm}.

\vspace{0.1cm}
\noindent
\textbf{Baselines: }
We choose \cite{fixmatch} as a baseline from the SSL domain, for interpretation. From the SSDG domain, we choose recent SOTA methods such as StyleMatch \cite{stylematch}, FBC-SA \cite{FBCSA}, and DGWM \cite{dgwm}. For a fair comparison, we only use work that has publicly available code bases, as it is imperative to use the same labeled/unlabeled data split for experiments since these models are highly sensitive to these splits.


\begin{figure}[t]
    \centering
    \begin{tabular}{cc}
        \fbox{\includegraphics[width=8.25cm]{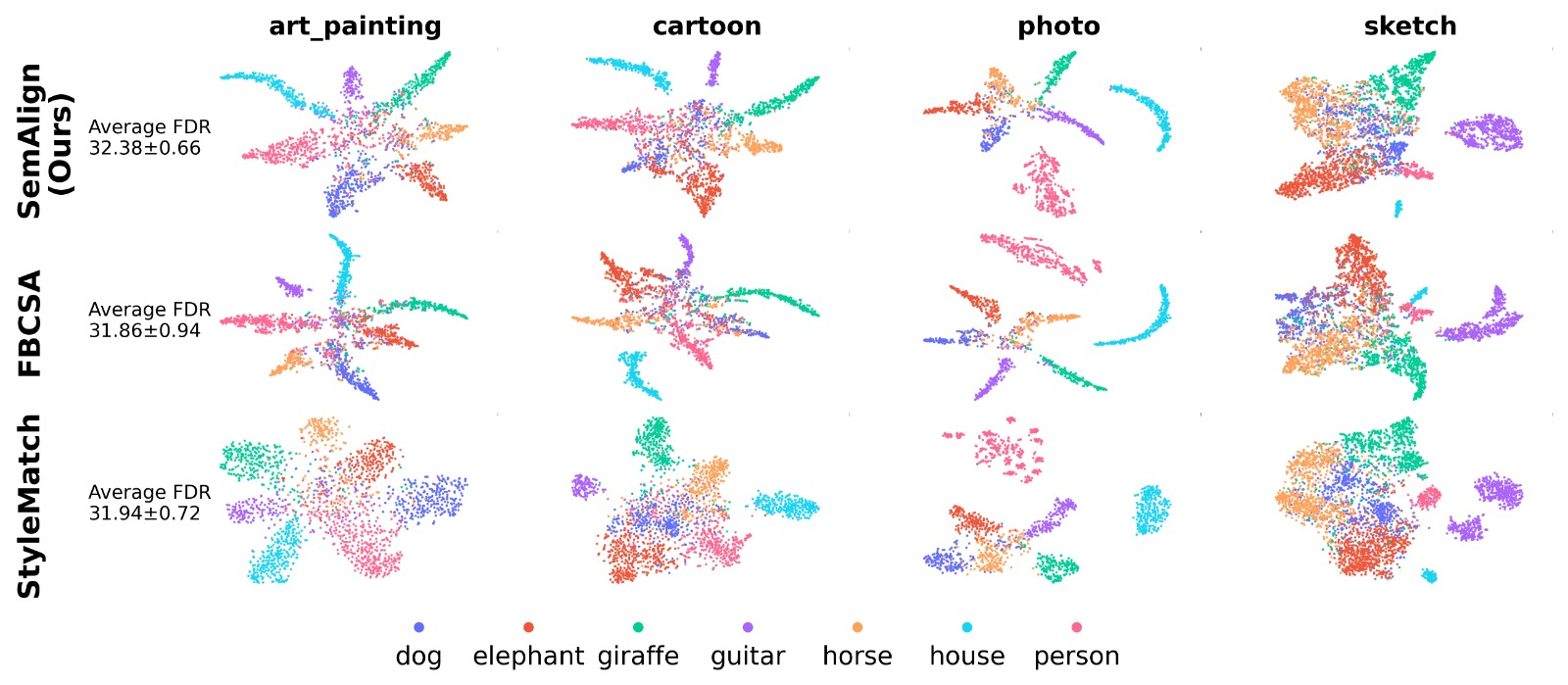}} &
        \fbox{\includegraphics[width=2.75cm]{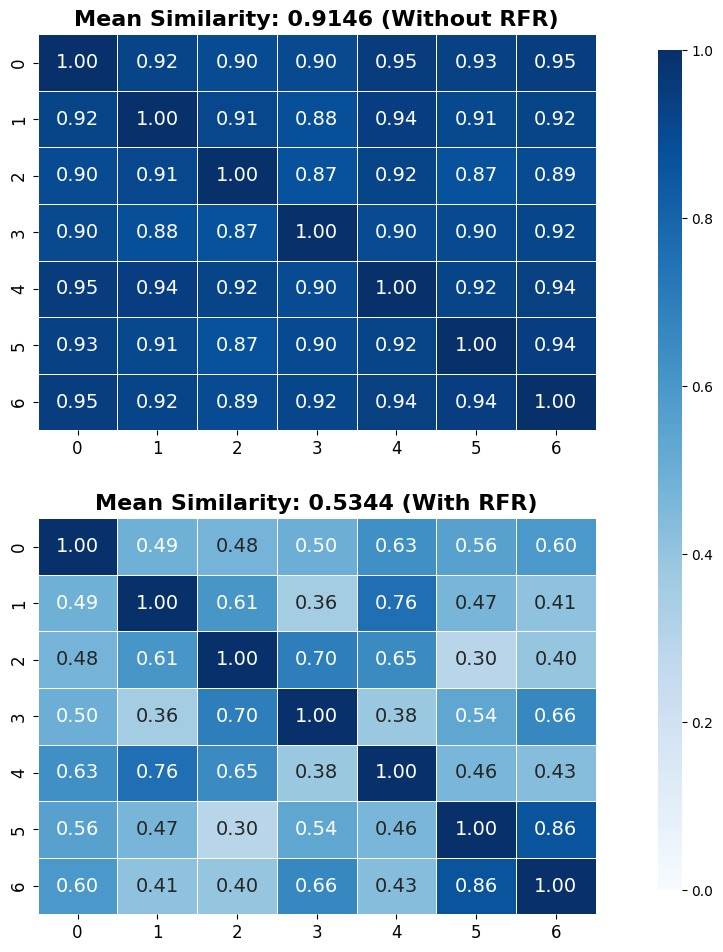}} \\
        (a) & (b)
    \end{tabular}
    \caption{(a) t-SNE visualization of the feature space for PACS dataset (10 labels per class). The average log Fisher Discriminant Ratio (FDR) ($\pm$1 standard deviation) was calculated over 5 random seeds, over all 4 domains. (b) Similarity matrices for the class prototype embeddings of the PACS dataset, before and after adding the RFR module. RFR reduces the similarity between class prototypes while preserving the semantics.}
    \label{fig:t-SNE}
\end{figure}

\vspace{0.1cm}
\noindent
\textbf{Model Generalization: }
For the settings with 5 and 10 labeled data samples per class, the model accuracies for the target domain for our method and the other baselines are reported in Tab. \ref{tab:ssdg_results}. As shown, our method achieves the best results across all experiments, while the second-best results are distributed among methods such as StyleMatch \cite{stylematch}, FBC-SA \cite{FBCSA}, and DGWM \cite{dgwm}. While identifying the second-best method is ambiguous due to varying performance across different datasets, our superiority is evident, as our model consistently leads across all the experiments by considerable margins. 


\vspace{0.1cm}
\noindent
\textbf{Effective Data Utilization: } Fig. \ref{fig:PL_Stats} shows the PL accuracy and the EDU for the PACS dataset for all combinations of source domains in the 10 labels per class setting. As shown, our PL accuracy is competitive with the other SSDG baselines, averaging just 0.45\% lower than the best method (FBC-SA). However, the EDU is significantly higher than other methods, exceeding FBCSA, which had the best PL accuracy, by an average of 5.25\% and StyleMatch, which has the second best EDU, by 2.00\%. In the absence of  $\mathcal{L}_\mathrm{out}$, the EDU of our method drops by 1.25\%. Additional EDU statistics are discussed in suppl.

\begin{figure}[t]
  \centering
   \includegraphics[width=0.5\linewidth]{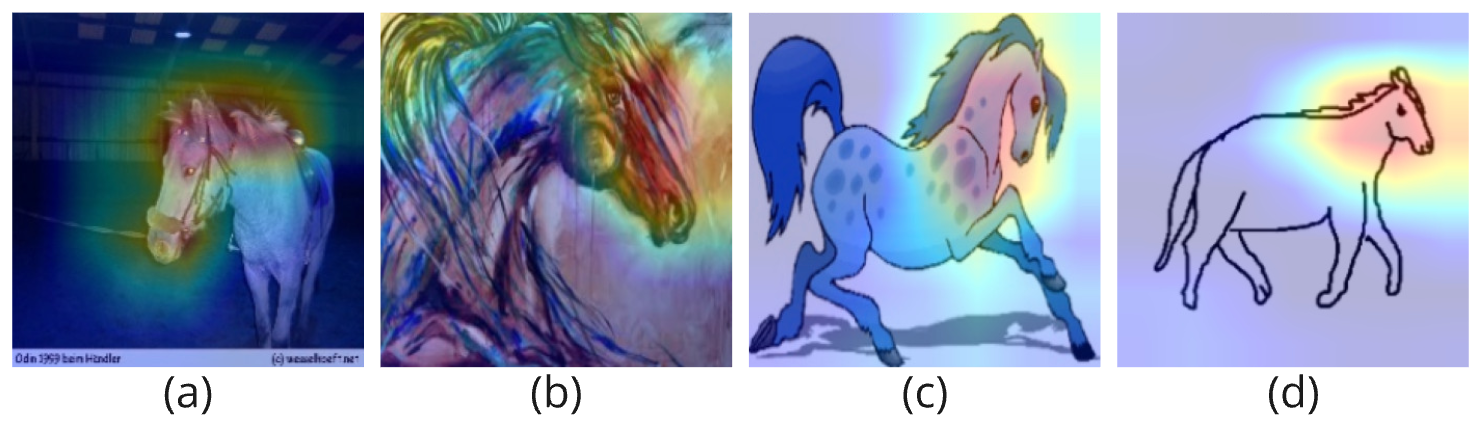}
   \caption{Grad-CAM visualization of model focus in (a) photo, (b) art-painting, (c) cartoon, and (d) sketch domains for horse class in PACS dataset. Our model consistently attends to the head of the horse and discriminates it from other classes.}
   \label{fig:gradcam}
\end{figure}



\vspace{0.1cm}
\noindent
\textbf{t-SNE Visualizations: } As shown in Fig. \ref{fig:t-SNE} our approach achieves a higher Fisher Discrimination Ratio (FDR) score, indicating that the features are more effective in distinguishing between classes. More visualizations can be found in suppl.

\vspace{0.1cm}
\noindent
\textbf{Grad-CAM Visualization: }As shown in Fig. \ref{fig:gradcam}, our model has learned invariant features across multiple domains, indicating good generalization capability.

\vspace{0.1cm}
\noindent
\textbf{Impact of Different Components: }\label{ablation_impact}
In Tab. \ref{tab:ablation_study} we report the contribution of each key component of our approach by sequentially adding them to FixMatch \cite{fixmatch}, as most components are complementary to each other. 

\vspace{0.1cm}
\noindent
\textbf{RFR Module:} Refinement of CLIP features using the RFR module and $\mathcal{L}_\mathrm{orth}$ gains 0.4\% improvement in the presence of $\mathcal{L}_\mathrm{SA}$ and gains 0.7\% in the presence of both $\mathcal{L}_\mathrm{SA}$ and $\mathcal{L}_\mathrm{con}$. 
As shown in Fig. \ref{fig:t-SNE} (b), this reduces the similarity between prototype embeddings, suggesting improved discriminability.

\vspace{0.1cm}
\noindent
\textbf{Feature Space Constraints: }Both FBC-SA \cite{FBCSA} and our method impose feature space consistency constraints. 
As demonstrated in the t-SNE plot, both the methods result in more discriminative and well-separated feature spaces.
However, our approach achieves a higher FDR score, showing its better discriminability.


%% file: sec/5_conclusion.tex
\section{Conclusion}
We present a novel approach that addresses the SSDG problem. We map intermediate features of the model to the semantically rich feature space of a VLM in a domain-invariant manner encouraging domain-invariant feature learning. Further, we introduce a set of simple data augmentation methods to encourage domain invariant feature learning and reduce overfitting. Our method is also capable of utilizing all data samples into the learning process. Results on four benchmarks, against four top-performing SSDG methods, show that our model has notable gains. A potential limitation of our method is the requirement of meaningful class labels which is left for future work.

\subsubsection{Acknowledgements} We would like to thank the Department of Electronic and Telecommunication Engineering, Univerity of Moratuwa, SriLanka for providing the required computational resources. Computational resources used in the project were funded by the National Research Council of Sri Lanka under the grant 19-080 and Accelerating Higher Education Expansion and Development (AHEAD)